\documentclass[sigconf,anonymous=false,nonacm=true]{acmart}
\usepackage{tikz}
\usetikzlibrary{bayesnet}
\usetikzlibrary{arrows}
\usetikzlibrary{backgrounds}
\usetikzlibrary{calc}
\usepackage{graphicx}
\usepackage{subcaption}
\usepackage{algorithm}
\usepackage{algorithmicx}
\usepackage[noend]{algpseudocode}
\usepackage{caption}
\usepackage{amsfonts}
\usepackage{amsmath}
\usepackage{hyperref}
\usepackage{booktabs}
\usepackage{multirow}

\AtBeginDocument{%
  \providecommand\BibTeX{{%
    \normalfont B\kern-0.5em{\scshape i\kern-0.25em b}\kern-0.8em\TeX}}}

\acmDOI{10.1145/nnnnnnn.nnnnnnn} 
\acmISBN{978-x-xxxx-xxxx-x/YY/MM} 
\acmConference[GECCO '24]{The Genetic and Evolutionary Computation Conference 2024}{July 14--18, 2024}{Melbourne, Australia}
\acmYear{2024}
\copyrightyear{2024}

\begin{document}

\title{Function Class Learning with Genetic Programming: Towards Explainable Meta Learning for Tumor Growth Functionals}

\author{E.M.C. Sijben}
\affiliation{%
 \institution{Centrum Wiskunde \& Informatica}
 \city{Amsterdam}
 \country{the Netherlands}
}
\email{evi.sijben@cwi.nl}

\author{J.C. Jansen }
\affiliation{%
 \institution{Leiden University Medical Center}
 \city{Leiden}
 \country{the Netherlands}
}
\email{j.c.jansen@lumc.nl}

\author{P.A.N. Bosman}
\affiliation{%
 \institution{Centrum Wiskunde \& Informatica}
 \city{Amsterdam}
 \country{the Netherlands}
}
\email{peter.bosman@cwi.nl}

\author{T. Alderliesten }
\affiliation{%
 \institution{Leiden University Medical Center}
 \city{Leiden}
 \country{the Netherlands}
}
\email{t.alderliesten@lumc.nl}
\renewcommand{\shortauthors}{Sijben et al.}




\begin{abstract}
Paragangliomas are rare, primarily slow-growing tumors for which the underlying growth pattern is unknown. Therefore, determining the best care for a patient is hard. Currently, if no significant tumor growth is observed, treatment is often delayed, as treatment itself is not without risk. However, by doing so, the risk of (irreversible) adverse effects due to tumor growth may increase. Being able to predict the growth accurately could assist in determining whether a patient will need treatment during their lifetime and, if so, the timing of this treatment. The aim of this work is to learn the general underlying growth pattern of paragangliomas from multiple tumor growth data sets, in which each data set contains a tumor’s volume over time. To do so, we propose a novel approach based on genetic programming to learn a \emph{function class}, i.e., a parameterized function that can be fit anew for each tumor. We do so in a unique, multi-modal, multi-objective fashion to find multiple potentially interesting function classes in a single run. We evaluate our approach on a synthetic and a real-world data set. By analyzing the resulting function classes, we can effectively explain the general patterns in the data.

\end{abstract}

\maketitle

\section{Introduction}
Paraganglioma are a slow-growing type of tumor. Predicting the growth of these tumors can be of great clinical help. 
They can cause serious complications, such as cranial nerve dysfunction and hearing loss. 
However, if the tumor stays small over the patient's lifetime, the likelihood of these complications is small. 
Therefore, treatment is often delayed until significant growth is detected or until the patient starts experiencing complaints. However, at that time, these complications might have become irreversible while they may have been avoided by treating the patient earlier.
Therefore, accurately predicting the growth of this type of tumor, possibly for the rest of the patient's lifetime, could be of great clinical value since it could support the decision of whether or not, and when, to give treatment. Previous work has fit known growth functions to tumor growth data ~\cite{heesterman2019mathematical}, but is limited in the number of functions considered. In this work,we provide a novel approach to learn an overall growth function from multiple data sets that can be specialized to each individual data set.

\begin{figure}[!t]
    \centering
\includegraphics[width=0.45\textwidth]{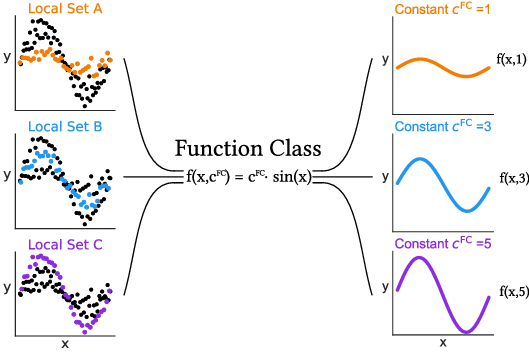}
    \caption{Visualization of function class learning. The global data set (black dots) consists of multiple local data sets (colored dots). A function class $f(x,c^{\mathrm{{FC}}}) = c^{\mathrm{{FC}}} \cdot \text{sin}(x)$ is learned that fits well with each local data set using a different value for function class constant $c^{\mathrm{{FC}}}$.}
    \label{fig:eye_catcher}
    \vspace{-1em}
\end{figure}

In machine learning, we often think of single data sets for which to learn a single model.
In practice, however, data may actually consist of multiple data sets, for instance, separate patients for which multiple measurements exist, physics experiments from different locations, or multiple measurements over time in general.
Similar mechanisms might underlie these different, yet related data sets.
We assume that this is also the case for the growth of paragangliomas: tumors that differ in characteristics, such as size and location, are assumed to exhibit a similar growth pattern or one of a limited number of potential patterns.
We refer to learning one or more general pattern from the \emph{global} data set, where the \emph{local} data sets have different constants pertaining to these patterns, as \emph{function class learning}.
We visualize this concept in \autoref{fig:eye_catcher}.

When combined with explainable AI techniques,
function class learning can be considered a unique way of performing explainable meta-learning. Understanding the overall patterns helps us understand the underlying system, while specific constants for specific scenarios, patients, or local data sets, help us make the actual predictions in general.
We argue that function class learning is more interpretable than creating a separate function for each local data set and more effective than creating a single (unparameterized) function for the global data set at once.
When learning a separate function for each local data set, we may find many (potentially overfit) different functions, which makes it difficult to see a general pattern.
Additionally, when training a separate function for each data set, there is no explicit drive to learn general overarching patterns.
When learning a single function for the global data set, it is harder to interpret the function's behavior per specific local data set. Furthermore, this most likely compromises performance on specific local data sets in favor of overall performance (imagine fitting a single sine function to the global data set that joins all local data sets).

\begin{figure}[!t]
    \centering
    \includegraphics[width=0.45\textwidth]{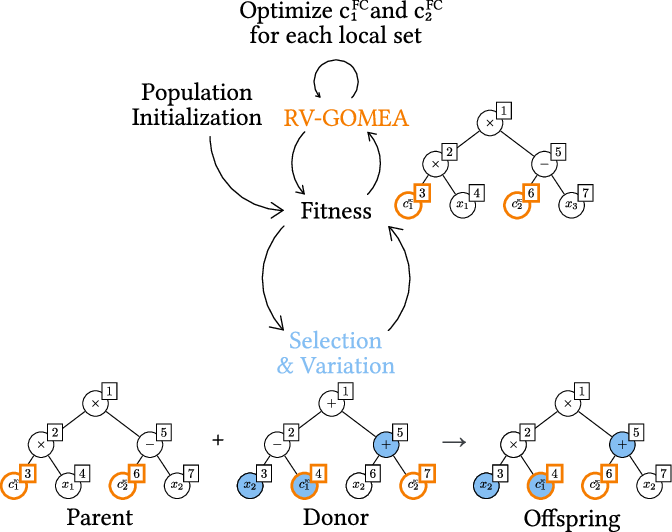}
    \caption{The Function Class GOMEA learning cycle. First, we initialize the population of function classes. Then, we calculate the fitness for each individual by tuning the function class constants (in orange) to each data subset by using RV-GOMEA. Next, we perform variation and selection in the typical optimal mixing way of GOMEA (illustrated in blue), and calculate the fitness again to test whether changes should be accepted. }
    \label{fig:training}
\end{figure}

In this work, we propose an algorithm for function class learning based on model-based evolutionary algorithms. Specifically, we use the Genetic Programming Gene-pool Optimal Mixing Evolutionary Algorithm (GP-GOMEA)~\cite{virgolin2021improving} to evolve the function class. Recent research suggests it is currently Pareto non-dominated  with respect to alternative symbolic regression algorithms in the trade-off between size and prediction accuracy of found expressions on the SR-Bench benchmark~\cite{la2021contemporary}, balancing between the most accurate algorithm and the one that delivers the smallest expressions. Since smaller functions are generally considered to have a higher chance of being interpretable, this property of GP-GOMEA is also of interest in this work. In GP-GOMEA, we introduce the Function Class Constant, $c^{\mathrm{FC}}$, as a new terminal type when performing symbolic regression. This terminal is optimized for each local data set separately during training, such that the function class is tuned to each local data set. In order to tune these $c^{\mathrm{FC}}$s, we use Real-Valued GOMEA (RV-GOMEA)~\cite{bouter2017exploiting} since it has been shown to be a powerful optimization tool for real-valued variables that is less prone to getting stuck in a local optimum than methods such as gradient descent. The general training cycle of the proposed algorithm is visualized in \autoref{fig:training}.
\vspace{-1em}
\section{Related work}
Most prior work acknowledges the slow-growing nature of paragangliomas but does not describe the underlying growth pattern~\cite{carlson2015natural,langerman2012natural,michalowska2017growth,prasad2014role}. 
Other work about paragangliomas in the head and neck area does hint somewhat at a certain growth pattern, but presents limited evidence~\cite{jansen2000estimation,heesterman2017age,heesterman2019mathematical}. In ~\cite{jansen2000estimation}, a higher percentage of tumors were observed to be growing when the tumor was of a size in the mid-range ($0.8-4.5 \text{cc}$). These observations could suggest a biphasic growth pattern, where the tumor growth first increases and later on decreases. However, they do not report about taking into account a possible selection bias that could be at play here since larger tumors are more likely to be treated. In ~\cite{heesterman2017age}, a model is introduced to estimate the chance of growth. However, in this work the model's predictive abilities are not tested. 
More closely to the work presented here, in~\cite{heesterman2019mathematical}, known growth expressions are fitted to tumor growth data sets. They conclude that s-shaped functions better fit the growth data. However, given their low number of measurements per tumor (three to fit the expression and none for validation or testing), a bias towards functions with a higher degree of freedom of the expression could be at play.
From this, we conclude that more research is needed to determine the underlying growth pattern of paragangliomas. We propose to do so by leveraging function class learning.

In function class learning, a \emph{functional} is learned, which is a function that returns a function. Most work on functionals in machine learning focuses on deriving functions from known functionals~\cite{cats2021machine, schmidt2021machine, dick2020machine}. 
Recent work on evolving symbolic density functionals, SyFes, uses a Genetic Algorithm (GA) to evolve functionals utilizing regularized evolution~\cite{ma2022evolving}. The mechanisms in SyFes work similarly to those in our approach; symbolic expressions are evolved with evolutionary algorithms, and the parameters in these expressions are fitted to related data sets. However, SyFes learns separate expressions for each of the related data sets. Interestingly, although they find functions of a known functional, their work does not focus on finding or analyzing new functionals. As a result, the better-performing functions presented in their work are studied separately and are not paired with a functional form, which we argue could hold unique overarching insight. 

Meta-learning is a term regularly used in machine learning ~\cite{vanschoren2019meta,spinelli2022meta,woznica2021towards}, and although it commonly refers to other types of methods than proposed in this work, there are some parallels. 
Meta-learning in machine learning often refers to learning
how to learn, such as learning how to learn to set the hyper-parameters or extract features. Hyper-parameter learning, for example, includes Neural Architecture Search (NAS) ~\cite{elsken2019neural}. NAS aims to optimize the structure of the neural network (e.g., with Evolutionary Algorithms (EAs)) and another optimizer to optimize the actual parameters of the network (e.g., gradient descent).
The main difference between these methods and our method is that we aim to learn meta-knowledge over related data sets, while meta-learning in learning hyper-parameters (or meta-parameters) aims to learn how to learn the model. 

In recent years, a connection between eXplainable Artificial Intelligence (XAI) and Symbolic Regression (SR) has been made in multiple works~\cite{evans2019s,ferreira2020applying}.
The main reason for this connection is that it is possible to evolve small expressions with SR, which may improve interpretability or knowledge discovery. 
In this paper, we propose a way to perform SR to learn over multiple related data sets rather than learning one model per data set. We do so utilizing GP-GOMEA, which, according to SR Bench, proved to be able to find relatively compact, yet accurate solutions. 
Additionally, we argue that function class learning can improve interpretability for related data sets, as it does not require interpreting one expression per local data set, but rather one function class per global data set. 

Multiple works combine GP with real-valued optimization to find constants~\cite{topchy2001faster,dick2020feature,kommenda2020parameter,koza1994genetic}.
The method presented in this paper performs constant optimization as well.
A key difference, however, is that we optimize a single SR expression, while we optimize constants anew for each local data set, using a novel fitness function.
This enables us to learn overarching function classes over the global data set, while also yielding solutions optimized for the local data sets.

\vspace{-0.5em}
\section{The GOMEA Family of EAs}
In this section, GOMEA, a family of model-based evolutionary algorithms is discussed. First, we explain the concept behind GOMEA. Then, we explain the specific variants of GOMEA that we use in Function Class GOMEA.
\vspace{-0.5em}
\subsection{Gene-pool Optimal Mixing Evolutionary Algorithm (GOMEA)}
GOMEA is a model-based EA that is effective in many domains such as discrete optimization, real-valued optimization, and GP~\cite{thierens2011optimal,luong2014multi,virgolin2021improving,bouter2017exploiting}. GOMEA differs from classic EAs in that it uses a linkage model that aims to capture the interdependencies within the genotype for a specific problem. This information is used during variation to prevent building blocks (or partial solutions) from being disrupted and to mix these blocks to create better solutions effectively. 

GOMEA uses a fixed-length string to represent the genotype such that a specific location in the string always refers to the same variable in the problem. In a so-called Family Of Subsets (FOS) knowledge model linkage information is captured, in the form of subsets of genes (string indices) that are assumed to be linked.

If the user knows linkage information a priori, they can provide it to GOMEA. Otherwise, it is learned from the population during evolution. 
To this end, Mutual Information (MI) is often used to measure linkage among gene pairs. A so-called Linkage Tree (LT) is built to represent variable dependence relations hierarchically. 
Computing the joint MI is costly. Therefore, GOMEA uses an algorithm to approximate joint MI called UPGMA~\cite{gronau2007optimal}.  

In GOMEA, every individual in the population undergoes variation every generation through Gene-pool Optimal Mixing (GOM) as described in \autoref{alg:GOM}. GOM uses the information in the FOS to replace linked genes at the same time. Suppose individual $\mathcal{P}_i$ undergoes GOM. First, $\mathcal{P}_i$ is cloned into offspring $\mathcal{O}_i$ (line 3). Then, each of the subsets in the FOS is considered in a random order (line 4). For each FOS subset, a donor is randomly selected (line 5). This selection can either be directly from the population, as with GP-GOMEA, or by sampling from a distribution learned from the population, as with RV-GOMEA. GOM replaces the values of the genes in $\mathcal{O}_i$ with the donor's genes, but only at the positions specified by the FOS subset (line 6). The change is kept if a replacement does not result in a worse fitness (lines 7-13).
\begin{algorithm}
\caption{: \textbf{GOM}(Individual $\mathcal{P}_i$, Population $\mathcal{P}$, FOS $\mathcal{F}$ )}
\label{alg:GOM}
\begin{algorithmic}[1]
\State $\mathcal{B}_i\gets \mathcal{P}_i$ \Comment{where $\mathcal{B}_i$ is a backup.}
\State $f_{\mathcal{B}_i} \gets f_{\mathcal{P}_i}$
\State $\mathcal{O}_i\gets \mathcal{P}_i$
\For{$\mathcal{F}_j \in \mathcal{F}$}
  \State $\mathcal{D} \gets$ RandomDonor$(\mathcal{P})$
  \State $\mathcal{O}_i \gets$ ReplaceAtIndices$(\mathcal{O}_i,\mathcal{D},\mathcal{F}_j$)
  \State $f_{\mathcal{O}_i} \gets$ ComputeFitness$(\mathcal{O}_i$)
  \If {$f_{\mathcal{O}_i} \leq f_{\mathcal{B}_i}$} \Comment{minimization is assumed here.}
    \State $\mathcal{B}_i \gets \mathcal{O}_i$
    \State $f_{\mathcal{B}_i} \gets f_{\mathcal{O}_i}$
  \Else 
    \State $\mathcal{O}_i \gets \mathcal{B}_i$
    \State $f_{\mathcal{O}_i} \gets f_{\mathcal{B}_i}$
  \EndIf
\EndFor
\end{algorithmic}
\end{algorithm}

After processing all FOS elements, $\mathcal{O}_i$ is added to the offspring set, and the subsequent population member is considered for GOM. After processing the entire population, the population is replaced by the offspring.

\subsection{GP-GOMEA}
GP-GOMEA~\cite{virgolin2021improving} is a variant of GOMEA used for genetic programming.
In GP-GOMEA, individuals are trees that adhere to a template with fixed node positions mapped to a fixed-length string. These trees contain operators and terminals, such as variables and Ephemeral Random Constants (ERCs). Thereby, the tree represents a symbolic expression. Applying GOM and learning the FOS can be done straightforwardly because fixed node positions are used. 

\vspace{-0.5em}
\subsection{RV-GOMEA}
RV-GOMEA~\cite{bouter2017exploiting} is a variant of GOMEA used for real-valued optimization. In RV-GOMEA, each gene represents a real-valued variable of the problem. Again, a FOS is built to learn the linkage of these genes. For each FOS element containing $k$ indices, a $k$-variate normal distribution is estimated using maximum likelihood (which only considers the best $35\%$ of the population). For the variation of an individual, the basic principle is to draw new elements from the previously learned normal distribution. 

Additionally, Forced Improvements (FI), Adaptive Variance Scaling (AVS), and Anticipated Mean Shift (AMS) are applied when generating new solutions. FI moves solutions out of a local minimum, AVS counterbalances the vanishing of variance as a result of selection, and AMS speeds up optimization on slope-like regions in the search space. More details can be found in~\cite{bouter2017exploiting}.

Additionally, in RV-GOMEA constraints can be defined which are prioritized over the fitness value. This allows to search constants for the local data set such that the constraints are not violated. 

\vspace{-0.5em}
\subsection{Multi-Modal GP-GOMEA}
Multi-Modal GP-GOMEA (MM-GP-GOMEA)~\cite{sijben2022multi} is a variant of GP-GOMEA.
In practice, when given the choice, a domain expert may prefer a model other than the one with the lowest training error for various reasons.
Therefore, MM-GP-GOMEA explicitly searches for multiple, diverse, models that trade-off different meanings of accuracy.

Learning a diverse set of solutions is achieved by implementing a multi-objective, multi-tree approach, i.e., each individual encodes not one but multiple trees. The two objectives in MM-GP-GOMEA are 1) the sum of each tree's Mean Squared Error (MSE) in a multi-tree and, 2) the diversified error, which is defined as the mean of the \textit{minimum} squared errors of the trees in a multi-tree. This multi-objective optimization approach finds an approximation front of models with low MSE's, yet increasingly, as the diversified error improves, with a different error distribution over the data points.

In this paper, we extend function class learning with this idea, as further elaborated in Section 4.

\section{Function Class GOMEA}\label{sec:fc}
In this section, we present our approach to function class learning, called Function Class GOMEA (FC-GOMEA).
We use a variant of GP-GOMEA to optimize the general structure of a function class. To specialize a GP-GOMEA solution in our framework (i.e., a function class) for each local data set separately, we introduce the function class constant, $c^{\text{FC}}$, as a possible terminal in the symbolic expressions. An $c^{\text{FC}}$ operates similarly to an ERC. However, an $c^{\text{FC}}$ does not have a specific value. Instead, it is optimized for each local data set separately with RV-GOMEA. A function class solution could for example be $c^{\text{FC}}_1 \times \text{x}_1 + c^{\text{FC}}_2$. 

\begin{figure}[!t]
    \centering
    \includegraphics[width=0.45\textwidth]{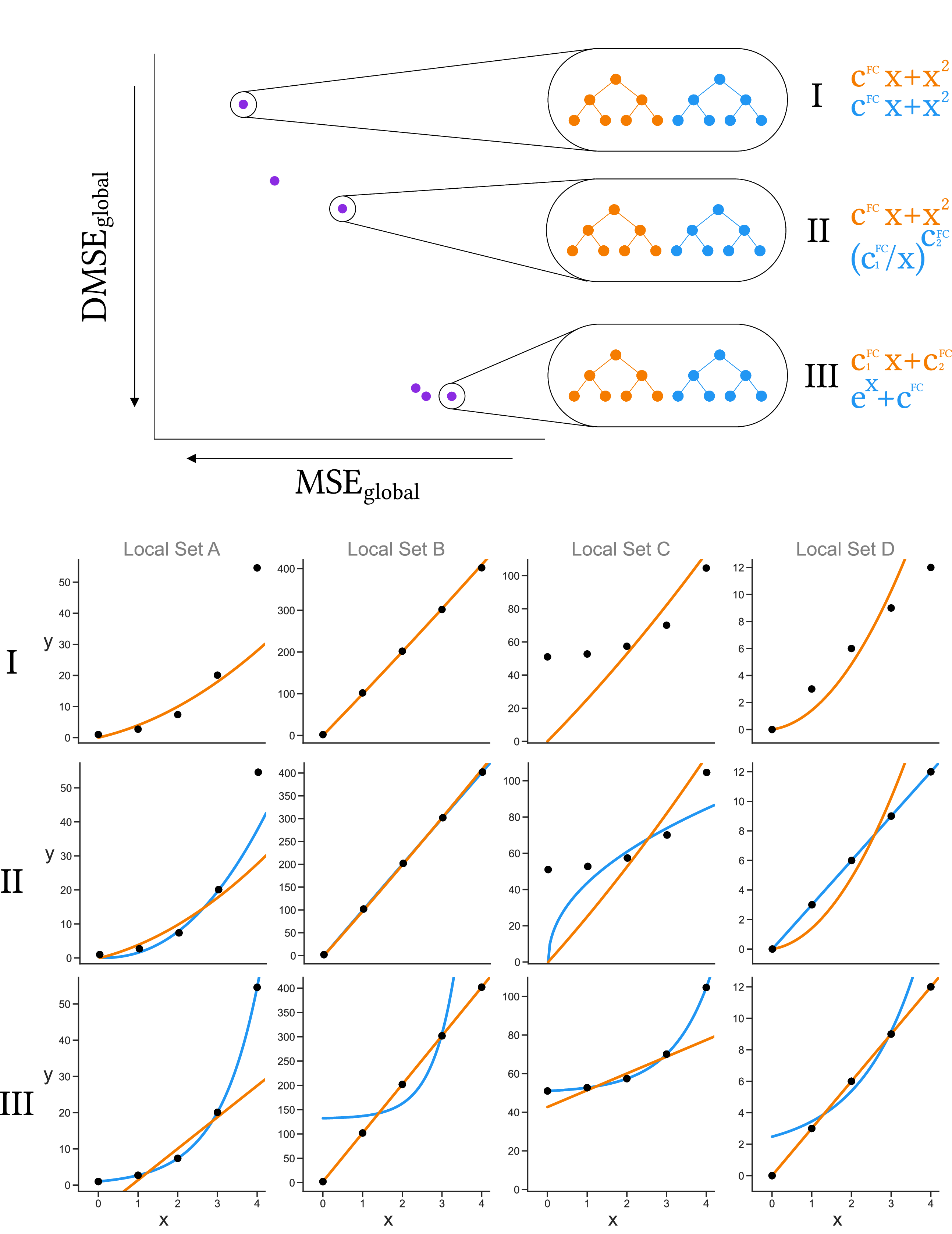}
    \caption{Visualisation of FC-GOMEA.
    Each local data set is either exponential or linear. The top scatter plot shows the approximation front with the trade-offs of the $\mathrm{MSE}_\mathrm{global}$ and the $\mathrm{DMSE}_\mathrm{global}$. Individual I has the lowest $\mathrm{MSE}_\mathrm{global}$ and thus the individual function classes fit the best on all local data sets, but there is no gain in using the classes together, because they are the same. Individual III has the lowest $\mathrm{DMSE}_\mathrm{global}$ and thus there is the most gain in using the function classes together. Individual II is somewhere in the middle: there is merit to using the function classes together, but at the same time, they fit relatively well on all local data sets. By utilizing multi-class learning we recover both function classes.}
    \label{fig:multiclass}
\end{figure}

To enable learning diverse sets of function classes, we use a specific variant of GP-GOMEA: MM-GP-GOMEA.
This allows us to learn a Pareto approximation front of sets of function classes that have different error distributions over the local data sets, optimizing for the error (MSE$_\text{global}$) and diversified error (DMSE$_\text{global}$). We visualize this concept in \autoref{fig:multiclass}. 
With this method, we recover both function classes that are equally plausible (lowest MSE$_\text{global}$), as well as function classes that specialize more towards a subset of local data sets (lowest DMSE$_\text{global}$), and options in between. This way the function classes in the multi-tree individual can, for example, depending on their objective values, be used independently, where the user or domain expert chooses the function class from a set of equally plausible function classes.  Or, together, where each function class is used for a specific local data sets. Additionally, a subset or the set of the multi-trees (and their function classes) can be used together as a way to express uncertainty about the actual underlying class. Here, one may take into account per local data set which of the function classes should be used per multi-tree. A similar approach to expressing uncertainty by means of multiple models is, for example, taken in weather forecasting.

\begin{algorithm}
\caption{: \textbf{FC-GOMEA}(population size $N$)}
\label{alg:FC}
\begin{algorithmic}[1]
\For{$i\in\{1,\dots,N\}$}
  \State $\mathcal{P}_i\gets$ CreateRandomSolution()
  \State EvaluateFCFitness($\mathcal{P}_i$)
\EndFor
\While{$\neg$TerminationCriteriaSatisfied}
    \State $\mathcal{F}\gets$ LearnLinkageModel($\mathcal{P}$)
  \For{$i\in\{1,\dots,N$\}}
    \State $\mathcal{O}_i\gets$ GOM($\mathcal{P}_i,\mathcal{P},\mathcal{F}$)
  \EndFor
  \State $\mathcal{P}\gets\mathcal{O}=\{\mathcal{O}_1,\dots,\mathcal{O}_{N}\}$
\EndWhile
\end{algorithmic}
\end{algorithm}
\autoref{alg:FC} shows the outline of the proposed method, which essentially follows the standard GOMEA outline but with a specialized fitness evaluation (see \autoref{alg:fitness}). We start by initializing the population, where each tree is built using variables, operators, and $c^{\text{FC}}$s (line 2). After initialization, we calculate the fitness for each solution in the population (line 3). We perform generations until one of the termination criteria is met (line 4). During each generation, we learn the FOS (line 5) and perform variation on each solution with GOM (lines 6 and 7). Finally, we replace the population with the offspring (line 8).

\subsection{Function Class Fitness}
\autoref{alg:fitness} shows how we calculate the function class fitness of an individual.

\begin{algorithm}
\caption{: \textbf{EvaluateFCFitness}(Individual $\mathcal{O}$,\newline Train input variables $\mathcal{X}_{\text{train}}$, Validation input variables $\mathcal{X}_{\text{val}}$, Train outcome variables $Y_{\text{train}}$, Validation outcome variables $Y_{\text{val}}$, Number of local data sets $M$)}
\label{alg:fitness}
\begin{algorithmic}[1]
\State $\hat{Y}\gets\text{Initialize}(\text{Size}(Y))$
\For{$m\in\{1,\dots,M\}$}
    \If{$c^{\text{FC}}{\text{Count}}(\mathcal{O}) = 0$}\Comment{if no $
c^\text{FC}$ nodes}
        \State $\hat{Y}_{\text{m}}\gets\text{Predict}(\mathcal{O}, \mathcal{X}_{\text{(val,m)}})$
    \Else
        \State  $\mathcal{O}\gets\text{RV-GOMEA}(\mathcal{O},\mathcal{X}_{\text{(train,m)}},Y_{\text{(train,m)}})$
        \State $\hat{Y}_{\text{m}}\gets\text{Predict}(\mathcal{O},\mathcal{X}_{\text{(val,m)}})$
    \EndIf
\EndFor
\State \Return $\text{MSE}_\text{global}(Y,\hat{Y})$
\end{algorithmic}
\end{algorithm}
\vspace{-0.5em}
We first initialize the array of prediction values $\hat{Y}$ (line 1).
Then, we loop over the $\mathcal{M}$ local data sets (line 2).
If the individual $\mathcal{O}$ does not have any $c^{\text{FC}}$s, we can immediately get its predictions $\hat{Y}_m$ on local data set $m$ (line 3-4).
Otherwise, we use RV-GOMEA to optimize the $c^{\text{FC}}$s on the train data and get the predictions on the validation data $\hat{Y}_m$ for local data set $m$ (lines 6-7).
Finally, we compute the fitness using $\text{MSE}_\text{global}$ (line 8). \\ We define the $\text{MSE}_\text{global}$ as follows,
\begin{align*}
    \text{MSE}_\text{global}(Y,\hat{Y}) &= \frac{1}{M}\sum_{m=1}^M \text{MSE}_\text{local}(Y_m,\hat{Y}_m),
\intertext{where,}
\vspace{-0.5em}
    \text{MSE}_\text{local}(Y_m,\hat{Y}_m) &= \frac{1}{N_m}{\sum_{i=1}^{N_m}{(Y_{i} - \hat{Y}_{i}
    )^2}},
\vspace{-0.5em}
\intertext{
where $N_m$ defines the number of records for local validation data set $m$. In FC-GOMEA, we take the sum of $\text{MSE}_\text{global}$ over the trees as error objective, such that,}\vspace{-0.25em}
    \text{MSE}_\text{global}(Y,[\hat{Y}^1,\ldots,\hat{Y}^T]) &= \sum_{t=1}^{T} \text{MSE}_\text{global}(Y,\hat{Y}^t),
\end{align*}
where $T$ is the number of trees and $\hat{Y}^t$ the predictions of tree $t$. For the diversified error, we take the mean over the minimum MSE for each local data set of the trees, such that, 
\vspace{-0.5em}
\begin{gather*}
    \text{DMSE}_\text{global}(Y,[\hat{Y}^1,\ldots,\hat{Y}^T]) = \\ \frac{1}{M}\sum_{m=1}^M \text{min}( \text{MSE}_\text{local}(Y_m,\hat{Y}_m^1),\ldots,\text{MSE}_\text{local}(Y_m,\hat{Y}_m^T)).
\end{gather*}
\vspace{-0.5em}

\subsection{Reducing Computational Cost}
The computational cost of a naive implementation of FC-GOMEA algorithm can be significantly reduced.
We implement six strategies to this end.

First, we implement a batching scheme to reduce the number of real-valued optimizations for evaluating the fitness. For each generation, we pick a different subset of local data sets that is used to evaluate for calculating the fitness. We keep track of the non-dominated solutions in an archive. This archive is emptied after each generation. Before emptying the archive, the solutions are evaluated on all local data sets, and the non-dominated solutions hereof are then stored in a second archive if they are non-dominated in that archive. Any solution in this second archive that is dominated by any newly added solution is deleted. This second archive is never emptied, and we output this archive to the user. 

Second, we implement a solution cache, which caches the fitness of expressions that have already been evaluated in the current generation. To identify equal expressions, we convert each expression to a string using infix notation and use it as the lookup key to the cache. If the key exists in the cache, we retrieve the already computed errors for that expression. Notice that we always calculate the MSE$_\text{global}$ before the DMSE$_\text{global}$, such that the errors can be re-used to calculate the DMSE$_\text{global}$. We use a hash map as the data structure for our solution cache. Since we use batching, the solution cache is emptied after each generation. 

Third, we implement caching for the output of sub-trees that do not contain $c^{\text{FC}}$s.
Therefore, when fitting the $c^{\text{FC}}$s to a local data set, we do not need to recompute parts independent of the $c^{\text{FC}}$s. 

Fourth, we identify sub-trees containing only terminals of the type $c^{\text{FC}}$ and replace them with a single $c^{\text{FC}}$ when optimizing constants. This is semantically equivalent and simplifies the expression, making the final expressions more readable.
Furthermore, simplifying the expression increases the chances of finding an equivalent expression in the solution cache.

Fifth, we parallelize Algorithm 1. We balance the workload to achieve a higher degree of parallelism by creating a priority queue such that expressions containing more $c^{\text{FC}}$s (which are assumed to take longer to optimize) are evaluated first. Threads are dynamically assigned work from the queue.

Sixth, we add a termination criteria to RV-GOMEA such that the algorithm is terminated once there has not been a significant improvement (at least $1\%$ decrease in fitness value) over the last four generations. 

Together, these optimizations significantly reduce the computational cost of our method, allowing many more generations to be evaluated within the allocated time.

\section{Evaluation: Tumor growth function class}
In this section the results of applying FC-GOMEA to the paraganglioma data are presented.
We have a real-world data set consisting of $226$ tumors with $n$ volume measurements per tumor (with $n \geq 4$) based on auto-segmentation as described in~\cite{sijben2024deep}. This data set contains $163$ tumors with $n \geq 5$.

We first apply FC-GOMEA to a synthetic data set that is based on the real-world data set, to confirm thecapabilities of FC-GOMEA to find ground-truth function classes. We then apply the FC-GOMEA to the real-world data set and analyze the results. 

\subsection{Generating Synthetic Data}
 In order to test the multi-class learning abilities of FC-GOMEA, we assume that there are not one but two different function classes representing the underlying growth function.
The first is the logistic function class: 
\vspace{-0.5em}
\begin{flalign*}
 \begin{array}{lcl}
  \text V^{\text{logistic}}(t,c1,c2,c3) &\!\!\!\! = \!\!\!\!& \frac{c_1}{1+ e^{-c_2\cdot(t-c_3)}}. 
 \end{array}
\end{flalign*}

In this equation, $c_1$ determines the maximum outcome of the function, $c_2$ determines the growth rate, $c_3$ is the inflection point, $V$ is the volume, and $t$ is the age. 
The second is the Gompertz function class: 
\vspace{-0.5em}
\begin{flalign*}
 \begin{array}{lcl}
  \text V^{\text{Gompertz}}(t,c_1,c_2,c_3)&\!\!\!\! = \!\!\!\!& 
  c_1\cdot e^{-c_2 \cdot e^{-c_3\cdot t}}. 
 \end{array}
\end{flalign*}
In this equation, $c_1$ determines the maximum outcome of the function, $c_2$ translates the curve in the direction of time, $c_3$ determines the growth rate, $V$ is the volume, and $t$ is the age. 

Both the logistic and Gompertz functions belong to the class of sigmoidal functions. Additionally, their constants serve similar purposes, despite their distinct equations. However, they differ in growth pattern since the logistic function class is symmetric and its inflection point occurs at half of the maximum ($c_1$), whereas the Gompertz is not symmetric, with an inflection point at $\frac{1}{e} \cdot c_1$. 

To generate the synthetic data set, we first fit the logistic and the Gompertz function class to the real-world data using RV-GOMEA for each of the 226 tumors. Notice that while we use RV-GOMEA both for generating the data as well as for fitting the function classes later on in FC-GOMEA, the input data is different each time (real-world data and synthetic data, respectively), such that this does not result in an unfair advantage. 
We fit the function class to the data using constraints formulated in consultation with a medical expert, such that the volume at birth is between 0 and 0.01 cc, and the volume at age 100 is less than 1500 cc, utilizing RV-GOMEA's built-in capabilities to define constraints.
Functions that do not adhere to these constraints are considered to be unrealistic~\cite{heesterman2016measurement}. We sample the fitted functions at the ages of the original data to represent the original data as closely as possible. In this, we alternate between the logistic function class and the Gompertz such that they are both equally represented in the final synthetic data set. We now have a data set that perfectly represents the two function classes. In order to test the effect of volume measurement variability on finding the ground-truth growth function classes, we construct two additional data sets, adding different degrees of noise to the synthetic data set. We add the noise such that bigger volumes also have a higher absolute level of noise, since this is more realistic. In Experiment I, we use GN$(0,0.05)$, i.e., Gaussian noise with a mean of 0, and a standard deviation of 0.05, and multiply it by the volume, as well as noise of GN$(0,0.15)$, again multiplied by the volume. 

\subsection{Algorithm Parameters and Settings}
We apply FC-GOMEA to both the synthetic and real-world tumor growth data set and use the settings in \autoref{tab:settings}. 
In this, $\text{n}_\text{RV}$ is the number of data points used for learning the $c^{\text{FC}}s$. Three data points are minimally needed for doing so, as a straight line can fit any two data points perfectly. We use points $\text{n}_\text{RV}+1,\ldots, \text{n}-1$  for validation, or in case  $\text{n}_\text{RV} + 1 =  \text{n}$, we use the $\text{n}^\text{th}$ point for validation (see \autoref{fig:outputs} for an illustration). Lastly, if $\text{n} > \text{n}_\text{RV}+1$, we use the n$^{\text{th}}$ point for testing. We only include local data sets (i.e., tumors) with at least $\text{n}_\text{RV}+1$ data points. So for example, if $\text{n}_\text{RV} = 4$ and $\text{n} = 5$ then the first 4 data points are used for fitting the function class, and the fifth data point is used for validation. But if $\text{n}_\text{RV} = 4$ and $\text{n} = 6$ then the first 4 data points are used for fitting the function class, the fifth data point is used for validation and the sixth data point is used for testing. 

Additionally, we define three constraints: 1) the predicted volume at birth can not be significant (i.e., $ 0 \text{ cc} \geq \hat{V} \leq 0.01 \text{ cc} $), 2) the predicted volume must be smaller than $1500 cc$ at the age of $100$ (we know the tumor will not grow infinitely), and 3) the function must be monotonically increasing.
In the algorithm, functions that violate fewer constraints are favored, even if they have a worse fitness, i.e. constrainted domination is used~\cite{deb2000mechanical}.

We have studied the known growth functions in ~\cite{heesterman2019mathematical} and converted them into tree shapes, and identified that none are full trees and that they all use a left-deep template, meaning each function always has a terminal as its right child. We have chosen to also use a left-deep tree template. Although this limits the search space and biases our search towards the known growth functions, we are still able to research the concept of function class learning within our computational budget. The left-deep template allows us to use a fairly large tree height while at the same time limiting the number of nodes per tree. In doing so, we can greatly reduce the time it takes to perform one generation for a specified tree height. Running FC-GOMEA with a full tree template of 32 nodes, and a real-valued evaluation budget of half a million, makes doing any relevant number of generations on the data sets in 120 hours infeasible. At the same time, using this left-deep template we are still able to express functions of a relevant intricateness (in terms of tree height).
Future work could focus on further reducing computational cost. This, possibly combined with a higher budget, could enable more generations with a full-template.

Since we are using a left-deep tree template, we include mirrored operators for the asymmetric operators (i.e., $\div$ and pow), denoted here by $\div_m$ and pow$_m$. Notice that, for example, it is impossible to express the function $\frac{c^{\text{FC}}_1}{x\cdot c^{\text{FC}}_2}$ otherwise, because the left-deep template combined with the standard implementation of division would not let us express the non-terminal tree $x\cdot c^{\text{FC}}_2$ in the denominator. 
Furthermore, the combination of operators e$^{(t_l\cdot t_r)}$, where the exponent of the left tree multiplied by the right tree (in our case, because of the left-deep template, always a terminal) is taken, commonly occurs in the known growth function classes. Therefore, we include this combined operator in our operator set as well. Notice that this can coincide with the pow operator when either $t_l$ or $t_r$ is of the type $c^{\text{FC}}$ since $a^x = (e^{ln(a)})^x = e^{ln(a)\cdot x}$ for $ln(a)>0$.

The number of real-valued evaluations per run of RV-GOMEA on a single local data set in FC-GOMEA is based on experiments with the synthetic data underlying different growth function classes (logistic and Gompertz), number of training data points and noise levels (see Supplementary). These experiments showed that the average MSE converged before one million evaluations. We notice that for most of them, the gains in average MSE was neglible after half a million evaluations. We thus set our evaluation limit to half a million evaluations. 

\begin{table}[t]
\caption{Algorithm parameter and experiment settings.}
\label{tab:settings}
\vspace*{-3mm}
\begin{tabular}{l|l|l}
\toprule
\textbf{Parameter} & \multicolumn{1}{c|}{\makebox[2.5cm]{\textbf{Experiment I}}} & \multicolumn{1}{c}{\makebox[2.5cm]{\textbf{Experiment II}}} \\
\midrule
Operators & \multicolumn{2}{c}{\makebox[5cm]{$+, -, \times, \div, \div_{m}, $ e$^{(t_l\cdot t_r)},$ $e^{t_l}$, $\text{pow}$, $\text{pow}_m$, log$_{p}$}} \\ \cline{1-3}
Terminals & \multicolumn{2}{c}{\makebox[5cm]{variable, $c^{\text{FC}}$}} \\ \cline{1-3}
Population size & \multicolumn{2}{c}{\makebox[5cm]{1,000}} \\ \cline{1-3}
Tree height & \multicolumn{2}{c}{\makebox[5cm]{4 ($9$ nodes)}} \\ \cline{1-3}
Tree template & \multicolumn{2}{c}{\makebox[5cm]{left-deep}} \\ 
\cline{1-3}
$\#$ trees & \multicolumn{2}{c}{\makebox[5cm]{2}} \\ \cline{1-3}
RV evaluations & \multicolumn{2}{c}{\makebox[5cm]{500,000}} \\ \cline{1-3}
$\#$ runs & \multicolumn{2}{c}{\makebox[5cm]{10}} \\ \cline{1-3}
$\#$ logical cores & \multicolumn{2}{c}{\makebox[5cm]{32}} \\ \cline{1-3}
Termination (hours) & \multicolumn{2}{c}{\makebox[5cm]{120}} \\ \cline{1-3}
Batch size & \makebox[2.5cm]{\{16,64\}} & \makebox[2.5cm]{16} \\ \cline{1-3}
n$_\text{RV}$ & \makebox[2.5cm]{\{3,4\}} & \makebox[2.5cm]{4} \\ \cline{1-3}
Data type & \makebox[2.5cm]{synthetic} & \makebox[2.5cm]{real world} \\ \cline{1-3}
Gaussian noise $\%$ & \makebox[2.5cm]{\{0,5,15\}} & \makebox[2.5cm]{not applicable} \\ 
\cline{1-3}
$c^{\text{FC}}$ values & \multicolumn{2}{c}{\makebox[5cm]{range(-10,000, 10,000)}} \\ 
\cline{1-3}
CPU architecture & \multicolumn{2}{c}{\makebox[5cm]{AMD EPYC 7H12}} \\ 
\bottomrule
\end{tabular}
\vspace{-1em}
\end{table}

\begin{figure}[!t]
    \centering
    \includegraphics[width=0.5\textwidth]{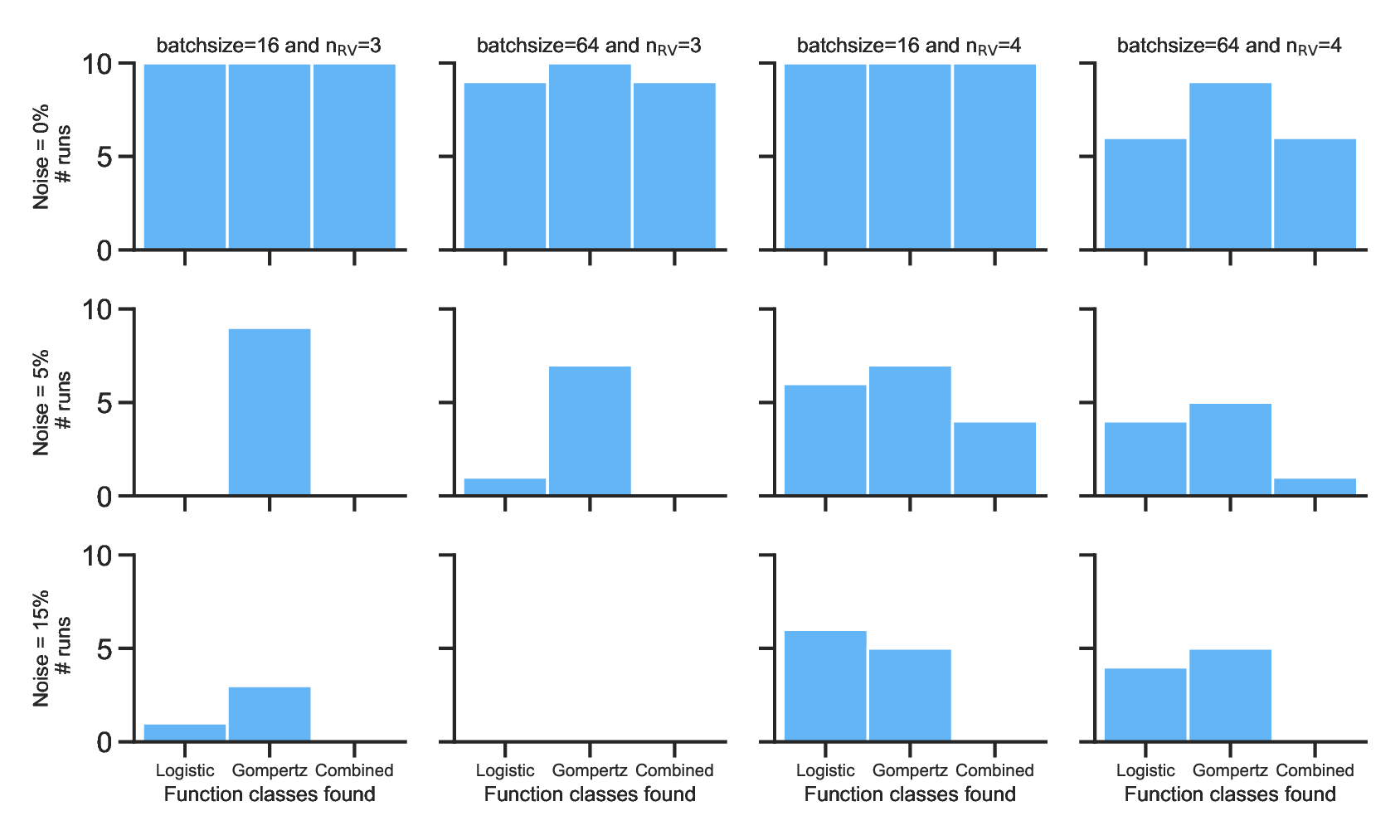}
    \caption{Histogram for number of times the correct function classes were found within any multi-tree of the full archive.   }
    \label{fig:countsGP}
    \vspace{-1em}
\end{figure}

\subsection{Experiment I: Synthetic Data}
\begin{figure}[t]
    \centering
    \includegraphics[width=0.5\textwidth]{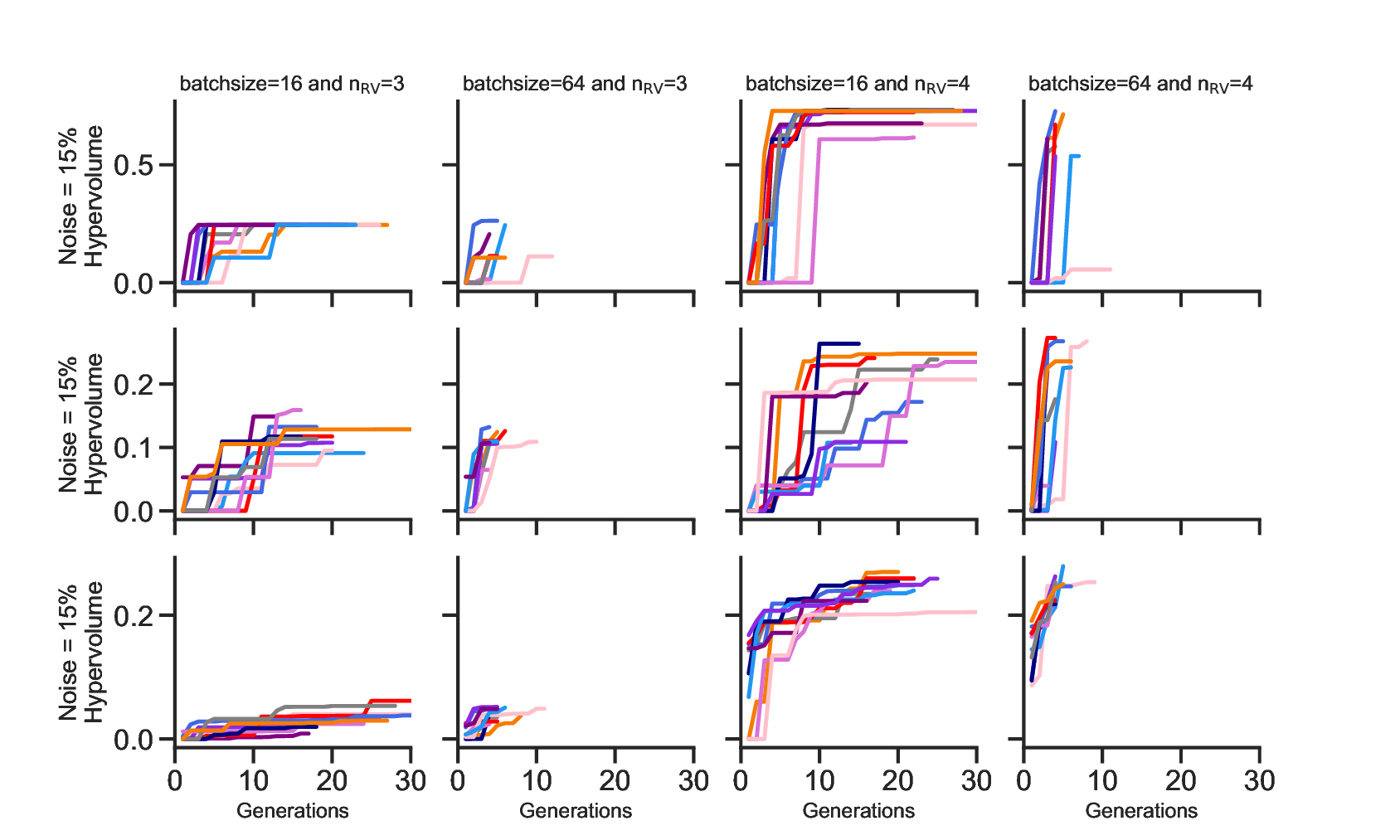}
    \vspace{-0.5em}
    \caption{Convergence plots for optimization using FC-GOMEA. Each color is a different run (different seed). It shows the HV of the global data set as function of the number of generations. Notice that if a line stops before 30 generations, it means run was terminated due to the time budget. In each row, the convergence for a different global data set is shown. The columns represent the different batchsizes and number of data points used for learning the $c^{\mathrm{FC}}\mathrm{s}$.  }
    \label{fig:convergenceGP}
    \vspace{-2em}
\end{figure}
The results of Experiment I are shown in \autoref{fig:countsGP} and \autoref{fig:convergenceGP}. \autoref{fig:countsGP} shows the number of runs (out of ten) in which the ground-truth function classes are found among all function classes in the elitist archive. \autoref{fig:convergenceGP} shows the convergence of running FC\allowbreak-GOMEA using different settings. The convergence is shown in terms of the Hyper Volume (HV)~\cite{zitzler1998multiobjective}. The HV is a measure of the volume covered by the approximation front w.r.t. a reference point. This point is calculated by first combining the fronts of all the runs with the same $\text{n}_{\text{RV}}$ and the same level of noise added to the data (i.e., the runs based on the same data), and by then taking the minimum and maximum values in each objective of the non-dominated solutions of this front. These points are then used to normalize the objective values in each front. Finally, we get the HV by computing the surface area covered by the front of a specific run with respect to reference point [1.0, 1.0]. 

When comparing the different settings in \autoref{fig:countsGP}, results improve when using $\text{n}_{\text{RV}} = 4$ instead of $\text{n}_{\text{RV}} = 3$, especially as the level of noise increases.
This makes sense, since if we only use $3$ data points for fitting the function class, any noise will have a major effect on the final fit.
Furthermore, we see that for $\text{n}_{\text{RV}} = 4$, the results always improves when using a batch size of $16$ compared to the batch size of $64$.
\autoref{fig:convergenceGP} indicates that a key reason for this is the time limit: the algorithm is terminated before it has converged.
For a batch size of $16$ however, we see that the runs seem to have converged.
Based on these experiments, we conclude that FC-GOMEA is indeed able to recover the intended function classes in many cases.

Especially for a batch size of $16$ and $\text{n}_{\text{RV}} = 4$, FC-GOMEA recovers the function classes in a significant number of cases for $0\%$ and $5\%$ noise.
For $15\%$ noise however, for none of the settings both function classes could be recovered correctly.
We argue that these results can be anticipated, considering that $15\%$ noise is quite significant, and the logistic and Gompertz function class share a lot of properties, since now the difference between the two function classes might be smaller than the noise for many of the local data sets. 
For both $5\%$ and $15\%$ noise the logistic and Gompertz function class are not generally at one of the extremes of the Pareto approximation front, and thus by performing function class learning in a single-modal fashion (even using the diversified error objective), would not be able to recover either of the function classes in most cases. In contrast, due to our multi-class approach we are still able to recover both of them a significant number of times.

We emphasise that the effect of the combination of measurement error and low amount of data on recovering the function class can be detrimental, but it can be mitigated by using the multi-class approach (at least to some extent).

\subsection{Experiment II: Real-World Data}
We compare the ten different runs of FC-GOMEA. We find that, although most runs performed quite well, compared to it, some runs underperformed (HV: $0.06$, $0.08$, $0.12$, $0.19$, $0.21$, $0.22$, $0.22$, $0.22$, $0.24$, $0.24$). We hypothesize that increasing the population size could improve consistency. However, here, we are interested in the best function classes for the paraganglioma use case and thus highlight the results of the best run. In the Pareto approximation front, we find four individuals containing altogether four distinct expressions, which we call FC1, FC2, FC3, and FC4. In the front, FC1 is combined with either itself or FC2, FC3, or FC4 in the multi-tree individuals.
The four expressions are as follows:
\vspace{-0.5em}
\begin{align*}
  \text{FC1}(t,c^{\text{FC}}_1, c^{\text{FC}}_2, c^{\text{FC}}_3) &= \frac{ c^{\text{FC}}_1}{e^{c^{\text{FC}}_2 \cdot t^{c^{\text{FC}}_3}}}\\
  \text{FC2}(t,c^{\text{FC}}_1, c^{\text{FC}}_2, c^{\text{FC}}_3)  &= c^{\text{FC}}_1\cdot e^{c^{\text{FC}}_2 \cdot e^{t\cdot\frac{t}{c^{\text{FC}}_3}}}\\
  \text{FC3}(t,c^{\text{FC}}_1, c^{\text{FC}}_2, c^{\text{FC}}_3, c^{\text{FC}}_4) &=  \frac{c^{\text{FC}}_1}{e^{c^{\text{FC}}_2 \cdot  e^{c^{\text{FC}}_3\cdot \frac{t}{c^{\text{FC}}_4}}}}\\
  \text{FC4}(t,c^{\text{FC}}_1, c^{\text{FC}}_2, c^{\text{FC}}_3, c^{\text{FC}}_4) &= c^{\text{FC}}_1 \cdot e^{c^{\text{FC}}_2 \cdot  e^{c^{\text{FC}}_3 \cdot \frac{t}{c^{\text{FC}}_4}}}
\end{align*}\vspace{-1em}

\begin{figure}[t]
\vspace{-1em}
    \centering
    \includegraphics[width=0.5\textwidth]{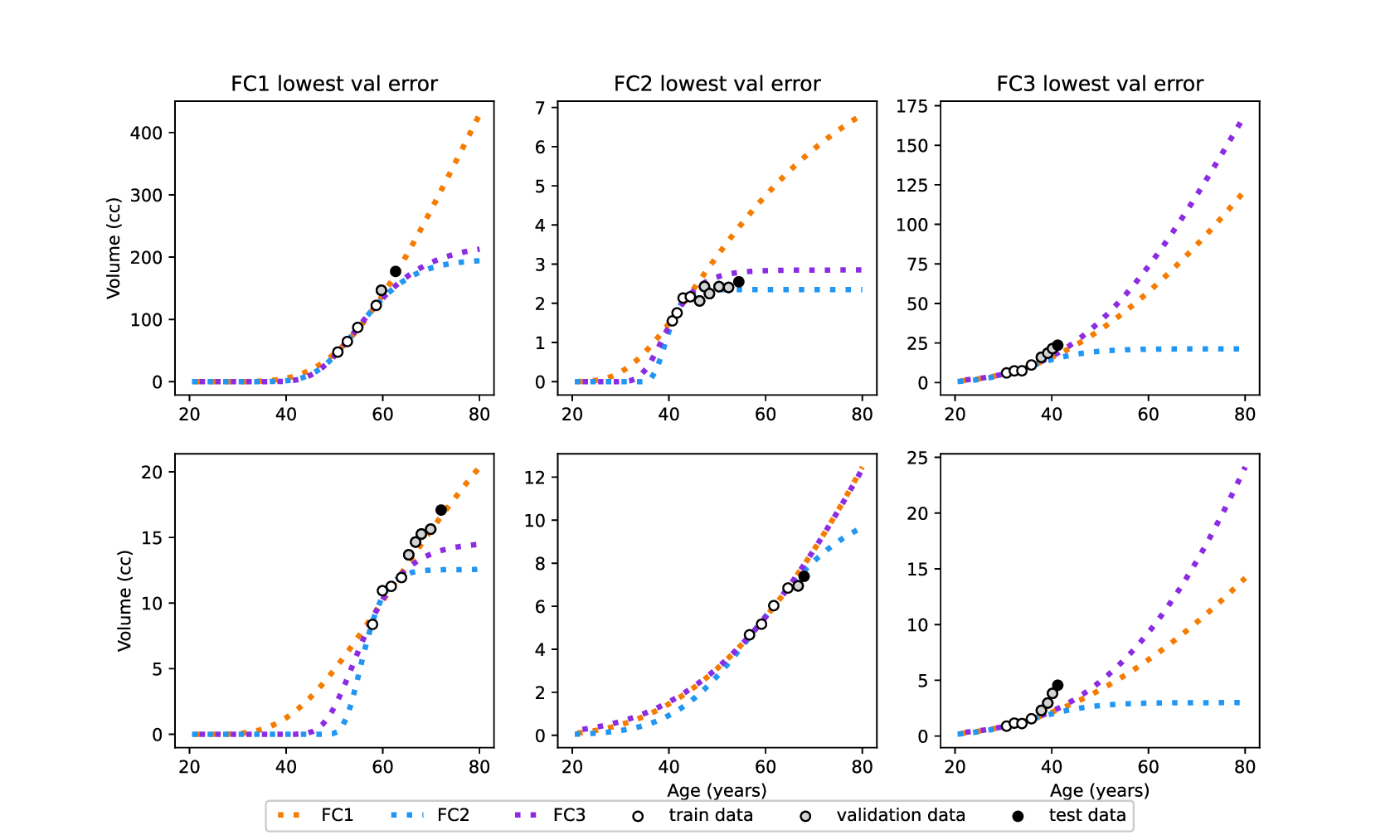}
    \caption{Predictions with the three function classes for actual growth data. Each column shows two examples in which that function class performed the best on the validation data of the tumor. Notice that the function class was only fitted to the train data.  }
    \label{fig:outputs}
    \vspace{-1em}
\end{figure}

\noindent FC3 and FC4 coincide since they are equivalent when flipping the sign of $c^{\text{FC}}_2$. The difference between these two functions in objective space is less than 0.01 \%. Thus, we only consider FC1, FC2, and FC3 now. Notice that FC3 coincides with the Gompertz function class, since $c^{\text{FC}}_3\cdot \frac{t}{c^{\text{FC}}_4} = \frac{c^{\text{FC}}_3}{c^{\text{FC}}_4} \cdot  t = c^{\text{FC}}_5 \cdot t$.
Although FC1 is similar to the Gompertz function class, it replaces $e^{c^{\text{FC}}_3\cdot t}$ by $t^{c^{\text{FC}}_3}$.
This replacement changes that part of the function from an exponential to a power component.
We speculate that this function class might better resemble the slow-growing nature of the paraganglioma.
FC2 shares many similarities with the Gompertz function class as well, but replaces $e^{c^{\text{FC}}_3\cdot t}$ by $e^{c^{\text{FC}}_3\cdot t^2}$.
We show predictions with these function classes in \autoref{fig:outputs}.

\section{Discussion}
When the measurements are more accurate, recovering the correct function class is more likely. This likelihood can be increased by searching for multiple classes in a multi-objective, multi-modal way. The resulting approximation front of classes provides options.

At least five volume measurements were needed to run FC\allowbreak-GOMEA on the paraganglioma use case. By only selecting the tumors with that amount of measurement and thus excluding tumors treated before the fifth measurement, selection bias could affect the results. Therefore, when using the function class, we must assume that (early) treated tumors adhere to the same function class as untreated tumors. Furthermore, the number of available volume measurements to fit the function class will impact the reliability of the found constants. Therefore, providing uncertainty estimates could be helpful in practice, especially in the case of < 4 measurements. In general, further research and analysis on the found function classes would be needed to explore a possible clinical implication. Possibly, in practice, the different function classes could be used as an ensemble representing not only a prediction but also an uncertainty estimate for the tumor growth. However, this again should be further researched. 

In order to run FC-GOMEA, we have used specific settings and parameters. These settings and parameters will affect the outcome and, thus, the found function classes. Choosing the optimal settings is a complex problem covering a whole field in itself. Considering that we recovered known function classes and new relevant function classes, we conclude that we get relevant outputs to this problem using these settings.

Although we present strategies to reduce the computational cost of the method presented in this paper, it still has a high cost. The cost grows, especially as the population size, tree height, and the number of RV evaluations increase. In comparison, tuning the constants of an already learned function class to a new data set takes only a fraction of the time. Additionally, predicting the outcome variables for data points with an instance of a function class is cheap, as it only requires evaluating a simple expression. So, the high costs apply primarily to the learning process and not the final prediction. Further efficiency enhancements may be possible by using less function evaluations for RV optimization.

\section{Conclusion}
This paper presented, for the first time, a method to learn multiple function classes over multiple related data sets rather than a single model per data set. This method was implemented as FC-GOMEA. Our experimental results showed that our method could find relevant solutions for real-world and synthetic tumor growth data sets. The presented solutions identified the over-arching patterns of the related data sets while fine-tuning predictions for each local data set. The solutions presented are at least as explainable as human-made growth functionals. 

\begin{acks}
This research was funded by the European Commission within the HORIZON Programme (TRUST AI Project, Contract No.: 952060).
Furthermore, we thank NWO for the Small Compute grant on the Dutch National Supercomputer Snellius.
\end{acks}

\bibliographystyle{ACM-Reference-Format}
\bibliography{sample-base}

\appendix
\end{document}